\documentclass[10pt,twocolumn,letterpaper]{article}

\usepackage[pagenumbers]{cvpr} 

\definecolor{cvprblue}{rgb}{0.21,0.49,0.74}
\usepackage[pagebackref,breaklinks,colorlinks,allcolors=cvprblue]{hyperref}

\usepackage{graphicx}
\usepackage{amsmath}
\usepackage{amssymb}
\usepackage{booktabs}
\usepackage{float}
\usepackage{comment}
\usepackage{multirow}
\usepackage{makecell}
\usepackage{enumitem}
\usepackage{makecell}
\usepackage[most]{tcolorbox} 

\usepackage[capitalize]{cleveref}
\crefname{section}{Sec.}{Secs.}
\Crefname{section}{Section}{Sections}
\Crefname{table}{Table}{Tables}
\crefname{table}{Tab.}{Tabs.}

\def\paperID{17477} 
\def\confName{CVPR}
\def\confYear{2025}

\newcommand\Rebute[1]{{#1}}

\begin{document}

\title{HiRQA: Hierarchical Ranking and
Quality Alignment for \\Opinion-Unaware Image Quality Assessment
\vspace{-3mm}
}
\author{Vaishnav Ramesh, Haining Wang, and Md Jahidul Islam\\
{\small RoboPI laboratory, Department of ECE, University of Florida, US} \\
\thanks{Accepted for Publication in Springer Journal of Machine Vision and Applications, May 2026.}
}
\maketitle

\begin{abstract}
Despite significant progress in no-reference image quality assessment (NR-IQA), dataset biases and reliance on subjective labels continue to hinder their generalization performance. We propose \textbf{HiRQA} ({Hi}erarchical {R}anking and {Q}uality {A}lignment), a self-supervised, opinion-unaware framework that offers a hierarchical, quality-aware embedding through a combination of ranking and contrastive learning. Unlike prior approaches that depend on pristine references or auxiliary modalities at inference time, HiRQA predicts quality scores using only the input image. We introduce a novel higher-order ranking loss that supervises quality predictions through relational ordering across distortion pairs, along with an embedding distance loss that enforces consistency between feature distances and perceptual differences. A training-time contrastive alignment loss, guided by structured textual prompts, further enhances the learned representation. Trained only on synthetic image distortions, HiRQA generalizes to authentic degradations, as demonstrated through comprehensive evaluations on various unseen distortions such as lens flare, haze, motion blur, and low-light conditions. For real-time deployment, we introduce \textbf{HiRQA-S}, a lightweight variant with an inference time of only 3.5\,ms per image. Extensive experiments across synthetic and authentic benchmarks validate HiRQA's competitive performance, strong generalization ability, and scalability. \Rebute{The HiRQA model and inference pipeline are available at: \url{https://github.com/uf-robopi/HiRQA}.
}
\end{abstract}

\section{Introduction}
\Rebute{Image Quality Assessment (IQA) includes computational methods for quantitatively evaluating the perceptual and statistical quality of images, aligning closely with human visual perception~\cite{wang2006modern,wang2002image,wang2021survey}. These metrics support a range of computer vision tasks, from content curation to model supervision, and complement related visual perception tasks such as salient object detection~\cite{song2025pushing,song2023rethinking,song2022improving,chen2025adapting} and data-driven pipelines~\cite{song2025sharpedge} that rely on training data quality. While reference-based IQA methods depend on pristine images, such references are rarely available in real-world scenarios, motivating the development of No-Reference IQA (NR-IQA). NR-IQA plays a critical role in practical applications such as mobile photography~\cite{fang2020cvpr}, real-time video streaming~\cite{lorenzzo_quality_videoframes}, surveillance applications~\cite{chen2025unveiling}, and image restoration pipelines~\cite{liang2021swinir,wang2021real,zhang2023mc,jin2025mb, zhang2018adversarial,zhang2021deep,zhang2022enhanced} that improve perceptual quality under diverse distortion conditions.}


\vspace{1mm}

Recent NR-IQA methods have achieved increasingly accurate predictions that align closely with subjective human opinion scores. However, these models often struggle to generalize to unseen distortions or new content, performing poorly on cross-dataset evaluation. Furthermore, collecting large-scale opinion-labeled IQA datasets is expensive, and the resulting scores are often subjective and inconsistent, limiting the robustness of supervised methods.

\begin{figure}[t]
    \centering  \includegraphics[width=0.95\linewidth]{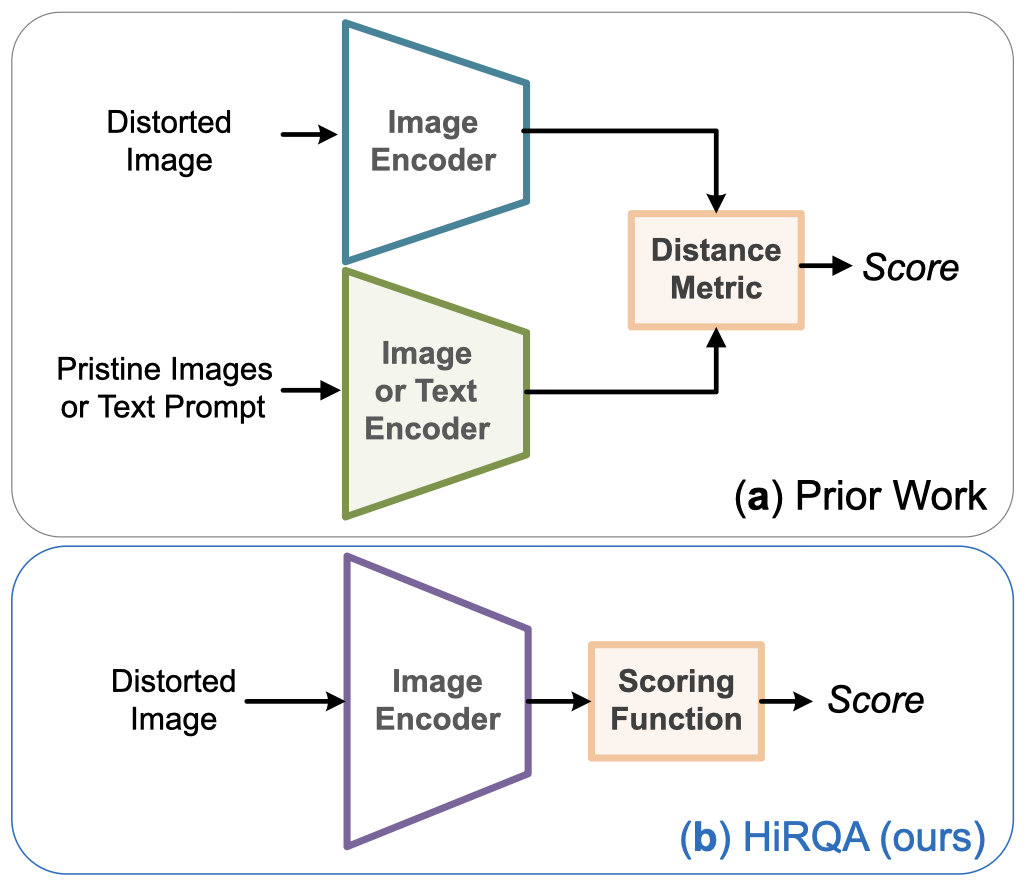}%
    \caption{Existing opinion-unaware IQA methods require additional modalities such as \textit{text prompts} or \textit{pristine} reference images for prediction. In contrast, HiRQA enables SOTA opinion-unaware IQA from the input image alone.}
    \label{Fig:Modelcomparison}
    \vspace{-1mm}
\end{figure}


\vspace{1mm}
To address these limitations, several Opinion-Unaware IQA (OU-IQA) methods have been proposed. These approaches eliminate the need for human-annotated labels and instead leverage handcrafted statistics~\cite{NIQE,ILNIQE} or self-supervised signals such as ranking consistency, distortion-specific cues, or large pretrained vision-language models (VLMs). As illustrated in Fig.~\ref{Fig:Modelcomparison}, many existing OU-IQA methods still rely on additional modalities such as a corpus of pristine images or textual prompts during inference. For instance, some methods~\cite{NIQE,ILNIQE,BabuOUIQAAuthentic} compute feature distances between a distorted image and a set of pristine natural images, but often fail to generalize to authentically distorted content. On the other hand, VLMs~\cite{wang2023exploring,Srinath_2024_WACV,agnolucci2024qualityaware} require handcrafted prompts and utilize large text encoders. {Although precomputing text embeddings avoids running a text encoder at deployment time, the approach still relies on the VLM's joint embedding space and prompt design. Consequently, backbone changes for deployment are nontrivial and typically requires large-scale pretraining, which complicates scaling to hardware constraints.} In contrast, we present a simpler approach that answers the following research question:

\begin{figure}[t]
    \centering
    \includegraphics[width=\linewidth]{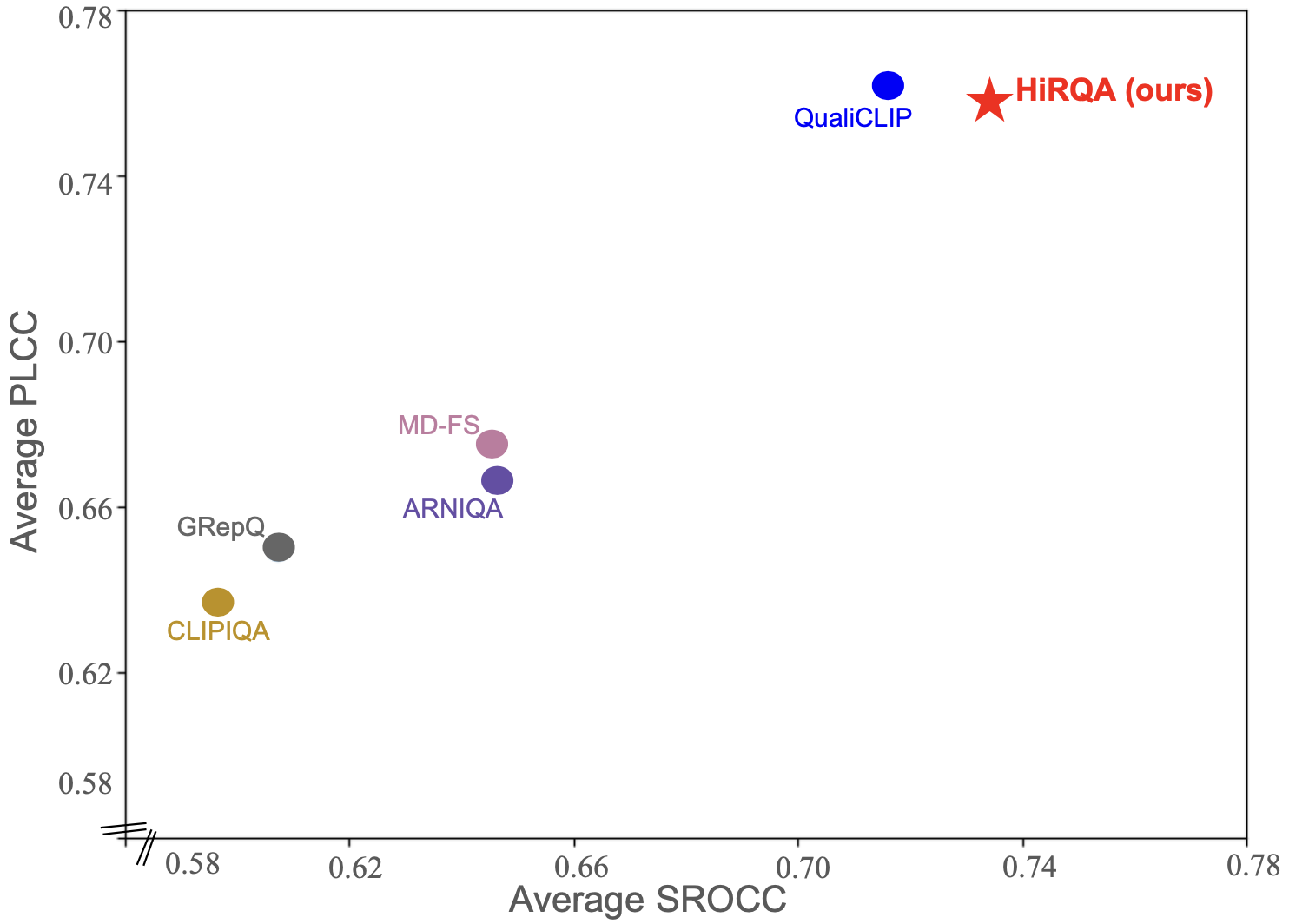}%
    \caption{Average SROCC vs PLCC performance across eight benchmark datasets. HiRQA achieves the highest correlation with human opinion, outperforming other SOTA OU-NRIQA methods.}%
    \label{fig:SROCCvsPLCC}
\end{figure}

\vspace{1mm}
\begin{quote}
   \textit{How to design an OU-IQA model that can learn to predict perceptual quality using only the input image without any reference images, text prompts, or other modalities at inference time?} 
\end{quote}

As shown in Fig.~\ref{Fig:Modelcomparison}, our proposed \textbf{HiRQA} model learns a hierarchical, quality-aware embedding space using synthetic distortions and relative quality supervision during training. Our training strategy integrates three complementary losses: (\textbf{i}) a novel pair-of-pairs ranking loss encourages the model to learn a globally consistent quality hierarchy by comparing relative distortion gaps across pairs of images; (\textbf{ii}) an embedding distance loss maps feature distances with perceived quality differences; and (\textbf{iii}) a text-image alignment loss applied only during training, uses structured distortion prompts within a CLIP-based contrastive setup to guide semantic representation. A comprehensive experimental evaluation of HiRQA on NRIQA benchmark datasets~\cite{sheikh2006statistical,larson2010most,ponomarenko2015image,lin2019kadid,ghadiyaram2015massive,hosu2020koniq,ying2020patches,fang2020cvpr} validates the effectiveness of our proposed learning strategy. As shown in Fig~\ref{fig:SROCCvsPLCC}, HiRQA outperforms most SOTA methods, achieving the highest correlation with human visual perception. {Additionally, in the supplementary material, we further evaluate its generalization to real-world distortions such as lens flare, motion blur, low-light, and atmospheric haze -- using density separation~\cite{ramesh2025dgiqadepthguidedfeatureattention}. We assess its performance on unseen enhancement datasets~\cite{dai2022flare7k,I-HAZE_2018,Chen2018Retinex,nah2017deep}, where HiRQA demonstrates a clearer delineation between high- and low-quality samples than state-of-the-art (SOTA) methods.}
\vspace{1mm}

\vspace{1mm}
\noindent
In summary, the main contributions of this paper are as follows:
\begin{enumerate}
\item We propose {HiRQA}, a novel opinion-unaware NR-IQA learning pipeline to predict image quality using a single distorted input, without relying on pristine reference images, opinion scores, or text prompts at inference time.  
\item We introduce a new loss formulation that combines a pair-of-pairs ranking loss and an embedding distance loss. This higher-order supervision strategy enables HiRQA to learn a hierarchical, distortion-aware feature space that aligns well with perceptual quality.
\item \Rebute{Extensive experiments on benchmark IQA datasets and detailed ablation studies validate the effectiveness of our proposed training strategy. HiRQA achieves SOTA performance on synthetic benchmarks 
and competitive performance on authentic datasets, while demonstrating 
promising generalization to unseen image distortions.}
\end{enumerate}

\noindent 
Additionally, we offer a lightweight variant, \textbf{HiRQA-S}, which employs a smaller backbone and achieves performance comparable to top-performing models, while having an inference speed of \textit{$3.5$} milliseconds per image.

\section{Background \& Related Work}

\noindent \textbf{Traditional Methods.} Non-Reference Image Quality Assessment (NR-IQA) traditionally relied on handcrafted features and statistical models to evaluate image quality without reference images, using subjective IQA databases as proxies for human perception. These methods often leveraged Natural Scene Statistics (NSS) and local descriptors to identify deviations indicative of distortions~\cite{gao2013universal,ghadiyaram2017perceptual,mittal2012no,moorthy2011blind,ye2012no,saad2012blind,xu2016blind,ye2012unsupervised}. While computationally efficient, they exhibited limited robustness when exposed to complex or unseen distortions and often struggled to generalize well. With the emergence of deep learning, Convolutional Neural Networks (CNNs) demonstrated improved performance by learning spatially hierarchical quality-aware features from large-scale data~\cite{kang2014convolutional,bosse2017deep,su2020blindly,zhang2018blind,zhu2020metaiqa,bianco2018use,lin2018hallucinated,talebi2018nima,jiang2021tongue}. \Rebute{ Despite these advances, CNN-based supervised methods face challenges in cross-dataset generalization, motivating efforts in continuous learning~\cite{zhang2022continual,zhang2024task} to improve robustness in diverse distortion conditions. Another line of work explores pairwise learning-to-rank as an alternative to direct quality regression. UNIQUE~\cite{zhang2021uncertainty}  supervises relative quality between image pairs via a fidelity loss, enabling joint training across IQA datasets without MOS alignment, while UNI-IQA~\cite{song2025uni} adopts a similar pairwise ranking strategy to unify natural and screen-content image datasets.} More recently, transformer-based models have surpassed CNNs by capturing global context and long-range dependencies, which are critical for perceptual quality understanding~\cite{you2021transformer,ke2021musiq,golestaneh2022no,yang2022maniqa}. Among these, MANIQA~\cite{yang2022maniqa} uses multi-dimensional attention mechanisms, achieving SOTA results across benchmark datasets.
\begin{figure*}[ht]
    \centering    \includegraphics[width= 0.95\linewidth]{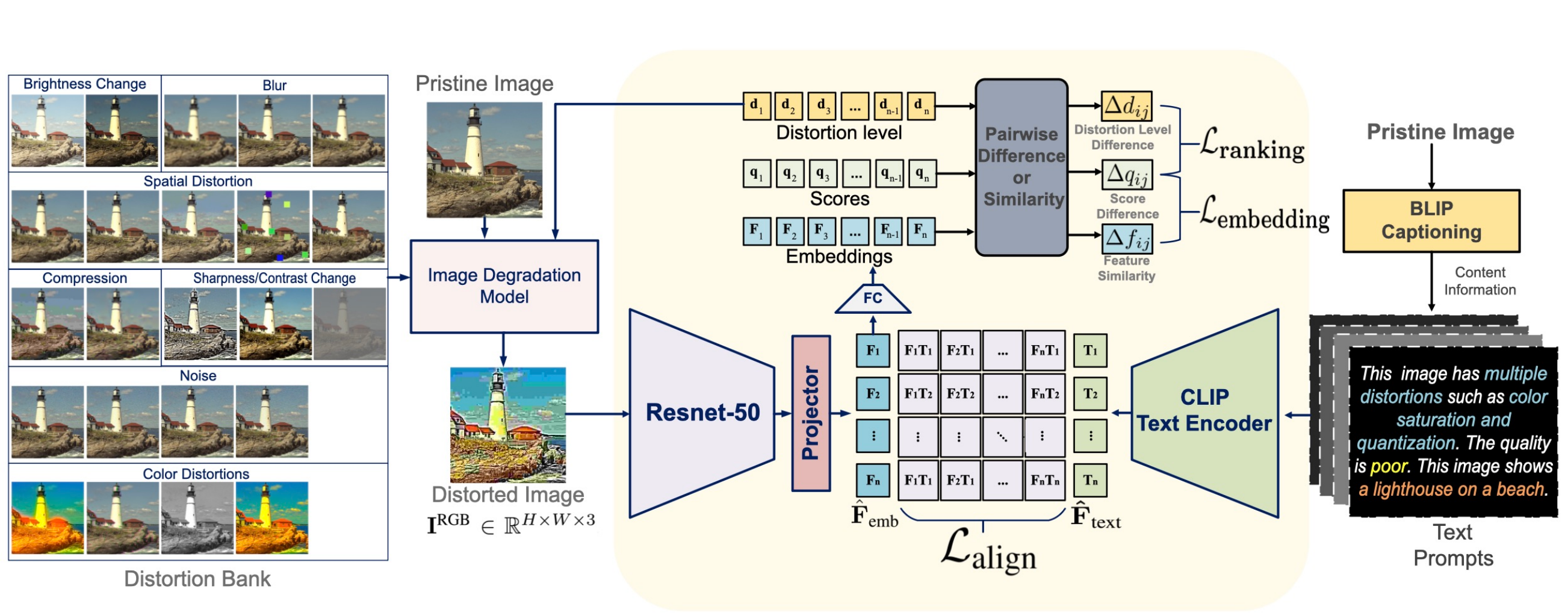}%
    \vspace{0mm}
    \caption{Training Pipeline of the proposed \textit{HiRQA}, a self-supervised Opinion-Unaware (OU) IQA framework. A ResNet-50 learns distortion-sensitive features from synthetically degraded images. These features are aligned with textual prompts where content information is generated using BLIP-based captioning and encoded via a frozen CLIP text encoder. The model is supervised using three losses: A higher-order pairwise ranking loss, an Image-text alignment loss, and an embedding distance loss capturing perceptual similarity. }
    \label{Fig:Model_Architecture}
    \vspace{-3mm}
\end{figure*}

\vspace{1mm}
\noindent
\textbf{Self-supervised Learning}. Self-supervised learning (SSL) has emerged as a compelling alternative for NR-IQA due to its ability to learn quality-aware representations without large-scale annotated datasets~\cite{agnolucci2024arniqa,saha2023re,madhusudana2021st,qpt,zhou2024visionlanguagemodelingcontent,Srinath_2024_WACV}. Motivated by the expensive and subjective nature of human opinion scores, SSL aims to leverage intrinsic properties of the data itself such as distortion patterns, ranking consistency, and augmentation invariance to supervise feature learning. Among various SSL strategies, contrastive learning has gained prominence for structuring the feature space to reflect perceptual similarity and distortion sensitivity by bringing similar (positive) pairs closer in the embedding space while pushing apart dissimilar (negative) pairs.

On the other hand, CONTRIQUE~\cite{madhusudana2021st} uses contrastive learning to align images with similar distortions while separating those with different types, effectively learning distortion-aware feature representations. REIQA~\cite{saha2023re} adopts MOCOv2~\cite{chen2020improvedbaselinesmomentumcontrastive} framework with dual encoders to separate content and quality features. {ARNIQA~\cite{agnolucci2024arniqa} employs the SimCLR~\cite{chen2020simpleframeworkcontrastivelearning} framework to align different images degraded with the same distortion configuration, allowing the model to learn a distortion-sensitive embedding space while minimizing content bias. }


\vspace{2mm}
\noindent
\textbf{Vision-Language Modeling}.
Textual representations provide rich, high-level semantic context that is inherently interpretable and transferable across tasks. This makes them powerful when paired with visual data. This synergy was first demonstrated by CLIP~\cite{radford2021learningtransferablevisualmodels}, which aligned images and text in a shared feature space via contrastive learning. By learning to associate visual patterns with natural language, CLIP enabled zero-shot generalization to a wide range of downstream tasks, inspiring a series of vision-language-based approaches for NR-IQA~\cite{wang2023exploring,agnolucci2024qualityaware,zhang2023liqe,zhou2024visionlanguagemodelingcontent,Srinath_2024_WACV,li2024boostingclipadaptationimage}.

\vspace{1mm}

CLIP-IQA~\cite{wang2023exploring} demonstrates that pre-trained vision-language models can assess image quality without supervision by measuring similarity between image embeddings and simple prompts such as \textit{“a good/bad photo”}. Building on this, later methods explored fine-tuning vision-language models to better align with human perceptual judgments. LIQE~\cite{zhang2023liqe} fine-tunes CLIP in a supervised multi-task setup that jointly predicts image quality, scene type, and distortion type. SLIQUE~\cite{zhou2024visionlanguagemodelingcontent}, trained on the TADAC dataset with text annotations describing content, distortions, and appearance, combines self-supervised learning with vision-language alignment through a dual-branch architecture.

Additionally, in Multimodal Large Language Models (MLLMs) for NR-IQA~\cite{wu2024comprehensive,you2024depictingscoresadvancingimage,wu2024qbenchbenchmarkgeneralpurposefoundation,wu2023qalign,chen2025groundingiqamultimodallanguagegrounding,chen2023iqagptimagequalityassessment,deqa_score}, image inputs are assessed via language-conditioned queries (e.g., “Rate the sharpness of this image”). While these methods enable flexible and instruction-based quality assessment, they require considerable computational resources due to the large model sizes and memory-intensive pipelines.

\vspace{2mm}

\noindent
\textbf{Opinion Unaware IQA.} 
While NR-IQA methods typically rely on subjective opinion scores for supervision, these scores are costly, inconsistent, and prone to dataset bias. OU-IQA aims to bypass this reliance by leveraging statistical regularities or synthetically generated distortions. Early OU-IQA models like NIQE~\cite{NIQE} and IL-NIQE~\cite{ILNIQE} employed hand-crafted statistical features, with IL-NIQE enhancing NIQE by incorporating quality-aware features such as Gabor filters and color gradients, comparing distorted image features to a corpus of pristine reference features. MDFS~\cite{ni2024opinionunawareblindimagequality} further extended this approach by extracting multiscale features and modeling them using multivariate Gaussian distribution. To better handle authentic distortions, CL-MI~\cite{BabuOUIQAAuthentic} proposed a self-supervised contrastive learning framework that explicitly disentangled content and quality information using a mutual information-based loss. However, despite these improvements, their performance remains limited on authentic datasets.

More recent methods leverage vision-language models~\cite{wang2023exploring,Srinath_2024_WACV,agnolucci2024qualityaware}. GrepQ~\cite{Srinath_2024_WACV} proposed a dual-encoder design, a fine-tuned CLIP encoder to capture high-level semantics, and a low-level encoder trained with a quality-aware contrastive loss using perceptual similarity functions as anchors. QualiCLIP\cite{agnolucci2024qualityaware} built on this by incorporating ranking losses and handcrafted antonymic prompts to align quality semantics with learned embeddings. {While effective, these approaches are tightly coupled to a specific pretrained encoder and its embedding space, which reduces architectural flexibility and limits scalability when adapting to different deployment constraints.}

\section{HiRQA Learning Pipeline}

\subsection{Model Architecture}
The proposed \textit{HiRQA} model aims to learn a distortion-aware and opinion-unaware hierarchical quality representation by leveraging images with known distortion levels. Given an input image $\mathbf{I}^\text{RGB} \in \mathbb{R}^{H \times W \times 3}$, it uses a ResNet-50~\cite{he2015deepresiduallearningimage} backbone to extract deep visual features, which are passed through an attention-based pooling module that projects them into a global embedding vector $\mathbf{F}_{\text{emb}} \in \mathbb{R}^d$. This vector is then mapped to a quality score $q \in [0, 1]$ through a fully connected decision layer. The complete end-to-end learning pipeline is illustrated in Fig.~\ref{Fig:Model_Architecture}

HiRQA's primary learning objective is a \textit{higher-order ranking loss}, $\mathcal{L}_{\text{ranking}}$, which enforces consistent relative ordering among input images based on their distortion severity. This ranking-based formulation enables the model to learn a perceptual scoring function without requiring subjective human labels. To further structure the learned feature space, we incorporate a contrastive \textit{image-text alignment loss},  $\mathcal{L}_{\text{align}}$~\cite{radford2021learningtransferablevisualmodels} to guide visual features using CLIP-based prompt embeddings (discarded at inference), and an \textit{embedding distance consistency loss}, $\mathcal{L}_{\text{embedding}}$ to encourage the geometry of the learned embedding space to reflect perceptual similarity. 

\subsection{Loss Functions}

\subsubsection{Higher-order Ranking Loss}

Ranking consistency is a popular self-supervised strategy in OU-IQA. Instead of using subjective opinion scores, the model learns from relative distortion severity among image pairs. dipIQ~\cite{ma2017dipiq} applies a RankNet-style pairwise ranking loss~\cite{ranknet}, while QualiCLIP~\cite{agnolucci2024qualityaware} uses a margin loss to rank degraded images of known distortion types.

We build upon the RankNet formulation by introducing the concept of higher-order pairwise ranking. Specifically, we compare the distortion and score differences between the two image pairs. If one pair exhibits a greater distortion gap than another, it should correspondingly show a larger difference in predicted quality scores. This \textit{higher order} comparison \textit{ensures a globally consistent and structured} ranking across the dataset. The effectiveness of this formulation is validated through ablation studies (Section~\ref{sec:ablation}), where it outperforms traditional ranking variants.  

Let $ q_i $ be the predicted quality scores and $ d_i $ the known distortion levels. From these, we construct all valid image pairs $ (i, j) $ where $ |d_i - d_j | > T_{\text{d}}$, $T_{\text{d}}$ is a threshold to filter out continuous distortion levels that are similar. Then we compute the absolute distortion and score differences as $ \Delta d_{ij} = |d_i - d_j| $ and $ \Delta q_{ij} = |q_i - q_j| $. Next, we compare these pairs, denoted $(i,j)$ and $(k,l)$, and define a binary label $y_{ij,kl}$ to indicate which pair has the greater distortion gap as follows:
\vspace{-1mm}
\begin{equation}
y_{ij,kl} =
\begin{cases}
1 & \text{if } \Delta d_{ij} > \Delta d_{kl} \\
0 & \text{otherwise}.
\end{cases}, \quad
\bar{y}_{ij,kl} = 1 - y_{ij,kl}
\label{eq:y_label}
\end{equation}


Each score difference $\Delta q$ is treated as a logit and supervised using binary cross-entropy (BCE). The pairwise ranking loss, $\mathcal{L}_{\text{r}}$ is defined over all pair-of-pairs $\mathcal{S}$ as:
\vspace{-1mm}
\begin{equation*}
    \text{BCE}(x, t) = -t \log \sigma(x) - (1 - t) \log(1 - \sigma(x))
\end{equation*}
\begin{equation*}
\mathcal{L}_{\text{r}} = 
\sum_{\mathcal{S}} {\displaystyle\frac{
\text{BCE}(\Delta q_{ij},\, y_{ij,kl}) +
\text{BCE}(\Delta q_{kl},\, \bar{y}_{ij,kl})}{2|\mathcal{S}|}
}
\label{eq:rank}
\end{equation*}
where $\sigma$ is the sigmoid function. Intuitively, when $y_{ij,kl} = 1$ (\textit{i.e.} $\Delta d_{ij} > \Delta d_{kl}$), the model is encouraged to predict a larger score gap $\Delta q_{ij}$ and a smaller gap $\Delta q_{kl}$ -- effectively aligning score differences with distortion severity. 

\Rebute{
We note that since $q_i \in [0,1]$, we have $\Delta q_{ij} \in [0,1]$ and $\sigma(\Delta q_{ij}) \in [0.5, 0.731]$, implying a theoretical floor for the ranking loss $\mathcal{L}_r$, meaning it cannot reach zero. However, the loss gradient:
\begin{equation*}
\frac{d\mathcal{L}_r}{d\Delta q_{ij}} = 
\sigma(\Delta q_{ij}) - t
\end{equation*}
remains correct in both directions: target $t=1$ gradients are always negative (increasing $\Delta q_{ij}$) and $t=0$ are always positive (decreasing $\Delta q_{ij}$). Furthermore, $t=0$ cases are better understood as regularization that prevents over-separation of perceptually similar images.}

Since we rank absolute score differences, the model may learn either MOS-like (higher score for lower distortion) or DMOS-like behavior. To \textit{enforce a consistent MOS} interpretation, we introduce a monotonicity regularization term $ \mathcal{L}_{\text{mreg}} $ that penalizes cases where the signed distortion and score differences are aligned, indicating that the model is incorrectly increasing the scores with distortion. Specifically, we minimize the following regularization objective:

\begin{equation*}
\mathcal{L}_{\text{mreg}} = \frac{1}{N} \sum_{i,j} \log\left(1 + e^{(q_i - q_j)(d_i - d_j)}\right).
\label{eq:mreg}
\vspace{-1mm}
\end{equation*}

The final higher-order ranking loss $\mathcal{L}_{\text{ranking}}$ combines both the terms:

\begin{equation}
\mathcal{L}_{\text{ranking}} = \mathcal{L}_{\text{r}} + \lambda_{\text{mreg}} \mathcal{L}_{\text{mreg}}.
\label{eq:ranking}
\end{equation}

\Rebute{In our implementation, HiRQA constructs ranking constraints within each batch and gradually learns a consistent ranking structure by comparing image pairs throughout training. Unlike margin-based ranking, which applies a fixed separation margin uniformly to all pairs, the pair-of-pairs formulation induces an adaptive margin from the batch itself. This provides finer-grained supervision and better preserves the relative structure of distortion severity.}

\subsubsection{Embedding Distance Consistency}
To reinforce a structured, quality-aware embedding space, we propose an \textit{embedding distance consistency} loss that formulates contrastive learning as a higher-order ranking problem. Unlike traditional contrastive learning approaches that rely on manually defined hard positive and negative pairs based on class labels or data augmentations, our formulation uses perceptual quality as the basis for supervision. Instead of explicitly specifying which samples are similar or dissimilar, we rank all image pairs according to their predicted quality differences. This allows us to learn a hierarchical representation of image quality, where the embedding space reflects perceptual similarity.

Let $ \hat{\mathbf{F}}_{\text{emb}}$ denote the $\ell_2$-normalized image embeddings and $ q_i $ the predicted quality scores. For each pair $ (i, j) $, the perceptual similarity $\Delta f_{ij}$ is computed as:
\vspace{-1mm}
\begin{equation*}
\Delta f_{ij} = \exp\left(\frac{\hat{\mathbf{F}}_{\text{emb},i} \cdot \hat{\mathbf{F}}_{\text{emb},j}^\mathbf{T}}{\tau_{\text{emb}}}\right)
\label{eq:distance}
\end{equation*}
where $ \tau_{\text{emb}} $ is a learnable temperature parameter. We then align the structure of this space with predicted quality score differences. Following our higher-order ranking framework, we form valid score-distance pairs $(i,j),(k,l)$  with $\Delta q_{ij}, \Delta q_{kl} > T_{\text{q}}$, where $ T_{\text{q}} $ is a threshold that filters out ambiguous or near-identical quality pairs.

This trick \textit{prevents noisy supervision} and encourages the model to focus on learning from meaningful perceptual differences rather than overfitting to minimal variations that may arise from prediction noise. With a label defined as $y_{ij,kl} = 1$ if $\Delta q_{ij} < \Delta q_{kl}$ (Eq.~\ref{eq:y_label}), the loss over all pairs of pairs ($\mathcal{S}$) is computed using binary cross-entropy by:
\vspace{-1mm}
\begin{equation*}
\mathcal{L}_{\text{ed}} = 
\sum_{\mathcal{S}} {\displaystyle\frac{
\text{BCE}(\Delta f_{ij},\, y_{ij,kl}) +
\text{BCE}(\Delta f_{kl},\, \bar{y}_{ij,kl})}{2|\mathcal{S}|}
}.
\label{eq:edist}
\end{equation*}

Moreover, to \textit{prevent embedding collapse} and promote diversity across feature dimensions, we introduce a covariance regularization term on the normalized embeddings $\mathbf{\hat{F}}_{\text{emb}}$. Specifically, we compute the sample covariance matrix $\Sigma$ and penalize redundancy by minimizing the squared Frobenius norm of its off-diagonal entries. The resulting regularization loss, $\mathcal{L}_{\text{cov}}$, is defined as:
\vspace{-1mm}
\begin{equation*}
\mathcal{L}_{\text{cov}} = \frac{1}{D^2} \sum_{i \ne j} \Sigma_{ij}^2, \quad 
\end{equation*}
\begin{equation*}
\Sigma = \frac{1}{N-1} (\mathbf{\hat{F}}_{\text{emb}} - \mathbf{\bar{F}})^\mathbf{T} (\mathbf{\hat{F}}_{\text{emb}} - \mathbf{\bar{F}})
\label{eq:cov}
\end{equation*}
where $\mathbf{\bar{F}}$ is the mean and $D$ is the embedding dimension. Then, the final embedding distance consistency loss is:
\vspace{-2mm}
\begin{equation}
\mathcal{L}_{\text{embedding}} = \mathcal{L}_{\text{ed}} + \lambda_{\text{cov}} \mathcal{L}_{\text{cov}}.
\label{eq:embed}
\end{equation}

\subsubsection{Image-Text Alignment Loss}\label{sec:img-text-align}
To further guide the structure of the quality-aware embedding space, we incorporate an auxiliary image-text alignment loss, inspired by the CLIP framework~\cite{radford2021learningtransferablevisualmodels}. We adopt this idea by pairing distorted images with descriptive quality prompts, which \textit{provide context about the type} of distortion, perceptual quality, and semantic content.

Prior models ~\cite{zhang2023liqe,wang2023exploring,zhou2024visionlanguagemodelingcontent} have demonstrated that using vision-language pretraining can enhance image quality representation. For instance, SLIQUE~\cite{zhou2024visionlanguagemodelingcontent} uses the TADAC
dataset, introduced in the same work, with quality-aware captions during training and discards the text encoder during inference. We follow this design: text prompts are only used at training time to shape the embedding space.

Let $\mathbf{\hat{F}}_{\text{text}}$ be the normalized embedding of the tokenized prompt from the CLIP text encoder. This loss adopts the standard cross-entropy formulation applied over similarity logits between image and text embeddings. It follows the InfoNCE loss~\cite{oord2019representationlearningcontrastivepredictive} computed as:
\vspace{-2mm}
\begin{equation*}
s_{i,j} =  e^{\tau_{align}} \, \mathbf{\hat{F}}_{\text{emb},i} \cdot \mathbf{\hat{F}}_{\text{text},j}^\mathbf{T}
\label{eq:contrastive_align}
\end{equation*}
\begin{equation}
\mathcal{L}_{\text{align}} = \frac{-1}{2N} \sum_{i=1}^{N}  
\log\!\left(
\frac{e^{s_{ii}}}{\sum_{j}e^{s_{ij}}}\cdot
\frac{e^{s_{ii}^{\mathbf{T}}}}{\sum_{j}e^{s_{ij}^{\mathbf{T}}}}
\right) 
\label{eq:contrastive}
\end{equation}

The text prompts used here during training are dynamically constructed and follow a consistent structure as follows:
\vspace{-1mm}

\begin{tcolorbox}[colback=black!95, coltext=white, boxrule=0pt, arc=4pt, left=4pt, right=4pt, top=4pt, bottom=4pt]
   {``This photo has \textcolor{cyan}{\textbf{\{a distortion/multiple distortions\} }} such as \textcolor{cyan}{\textbf{\{distortion names\}}}. The quality is \textcolor{yellow}{\textbf{\{quality adjective\}}}. This image shows \textcolor{orange}{\{semantic caption\}}.''}
\end{tcolorbox}
\vspace{1mm}

\Rebute{ Each prompt integrates three components. First, a \textit{distortion clause} describes either a single or multiple distortions. Second, a \textit{quality-level adjective} is assigned based on the distortion severity, drawn from the set: \{\texttt{excellent, good, average, poor, bad}\}, by rounding the continuous distortion level to the nearest integer. Notably, this mapping is a deterministic conversion of a known distortion parameter into a natural language, independent of any human opinion scores, thus preserving the opinion-unaware nature of the framework.} Finally, a \textit{semantic caption} is appended to describe the image content, which is automatically generated using a BLIP-based captioning model~\cite{li2022blip,li-etal-2023-lavis}. This structured prompting mechanism enables the model to \textit{correlate perceptual degradation} with both distortion types as well as the semantic context.

\begin{figure}[b]
    \centering \includegraphics[width=\linewidth]{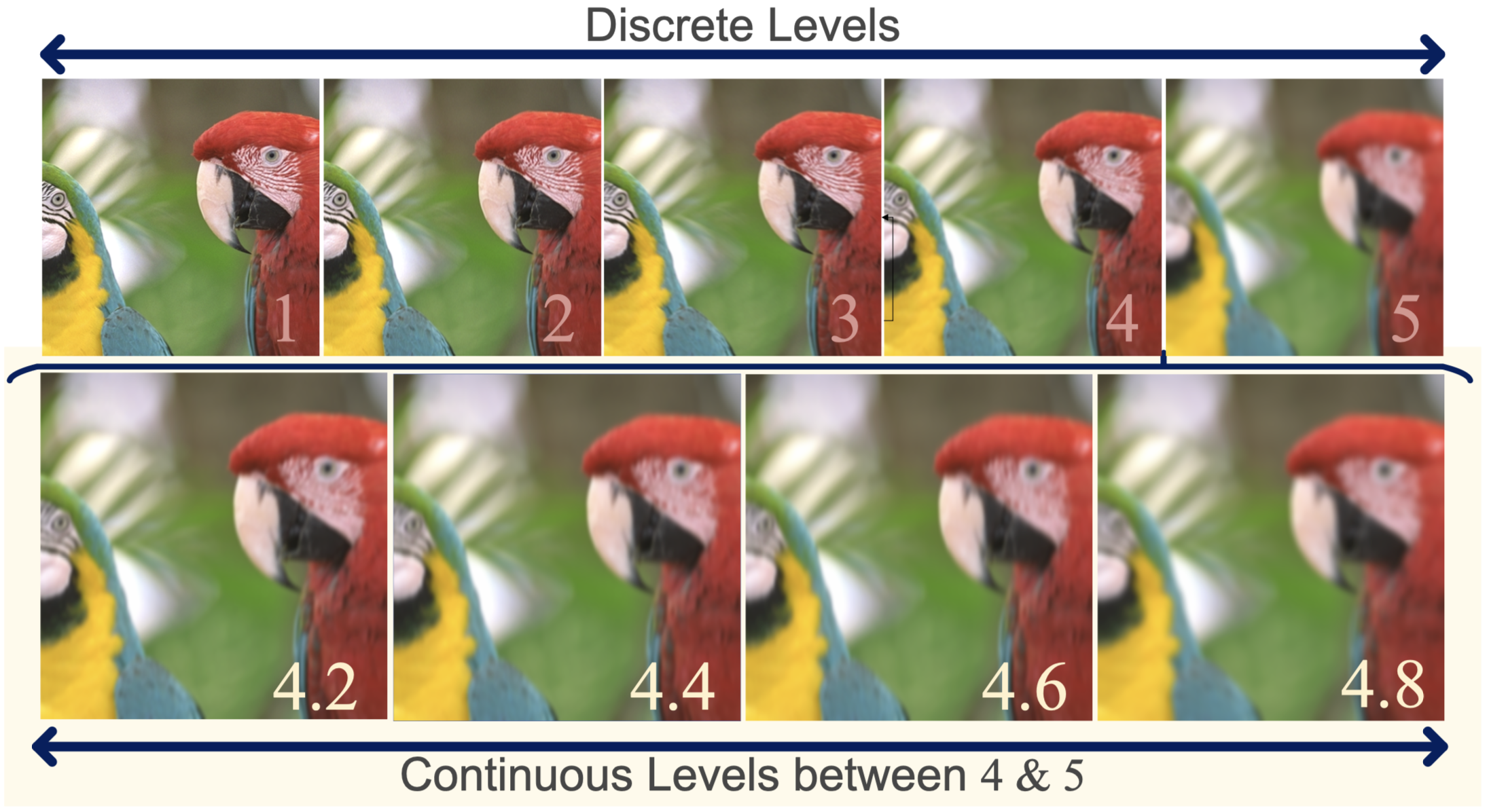}%
    \caption{Illustration of continuous distortion levels used in our training. The top row shows five discrete levels 1-5 for a particular distortion (Motion Blur), while the bottom row demonstrates continuous levels between levels 4 and 5.}%
    \label{fig: Image degradation model}
\end{figure}

\subsection{Image Degradation Model}\label{sec:idm}

Self-supervised NR-IQA methods commonly rely on synthetically degraded images for training. However, training solely on synthetic images limits model generalization to real-world distortions. Prior approaches, such as Re-IQA~\cite{saha2023re} and CONTRIQUE~\cite{madhusudana2021st} improve generalization by combining synthetic and authentic images. Moreover, models like QPT~\cite{qpt} and ARNIQA~\cite{agnolucci2024arniqa} have demonstrated that applying multiple simultaneous distortions to the same image simulates authentic degradations, boosting generalization.

Our degradation model follows a similar strategy to ARNIQA~\cite{agnolucci2024arniqa}. Specifically, we utilize 23 distortion types with five intensity levels as defined in the Kadid10k~\cite{lin2019kadid} dataset, spanning seven categories: brightness change, blur, spatial distortion, color distortion, compression artifacts, noise, and sharpness/contrast change. \Rebute{Notably, Kadid10k's five-level severity scale was designed to represent comparable perceptual degradation across distortion types, enabling consistent ranking supervision across different distortion types.}

For each training sample, we randomly select one or more distortion categories and sequentially apply multiple distortions up to a maximum number $L_\mathrm{dist}$. \Rebute{For images with multiple distortions, the overall severity is defined as the maximum distortion level among all applied distortions. This serves as a practical approximation motivated by the fact that when multiple degradations coexist, a more severe distortion may dominate visual perception and reduce the perceptual impact of milder ones.} In addition to this standard practice, we introduce the concept of ``\textbf{continuous distortion variation}" by interpolating the intensity between discrete levels, rather than using only fixed levels. This enables us to generate images with subtle quality differences as shown in Fig.~\ref{fig: Image degradation model}.

{We utilize the $140$\,K pristine images from the KADIS-700K dataset~\cite{deepfl-iqa}. We extract random crops from each image and apply degradations to generate ranked training images for self-supervised learning. For each crop, we first sample one or more distortion categories and then randomly select a single distortion type from each chosen category, such that we never apply two distortions from the same category within one sample. The selected distortions are sequentially applied up to a maximum of $L_\mathrm{dist}$, and their base discrete intensity levels $l$ are sampled following ARNIQA~\cite{agnolucci2024arniqa}. This enables a large combinatorial space of degradations, which produces diverse training samples.}

{Crucially, this strategy also provides a controllable supervision signal; for each crop we generate multiple degraded variants with different overall severities and use their known \underline{relative ordering} as rank supervision. To introduce continuous variation, we sample a small offset $\delta$ from a zero-mean Gaussian: $\delta \sim \mathcal{N}(0, \sigma_\text{off}^2)$, which is added to the base level and then clipped to the valid range $\{1,5\}$. The final intensity $l^* = l + \delta$ is used to linearly interpolate between the nearest discrete levels, generating subtle quality differences within the same nominal distortion category. This augmentation yields a finer spectrum of perceptual qualities, allowing the model to learn from both large and subtle differences.}


\begin{figure}[b]
    \centering \includegraphics[width=\linewidth]{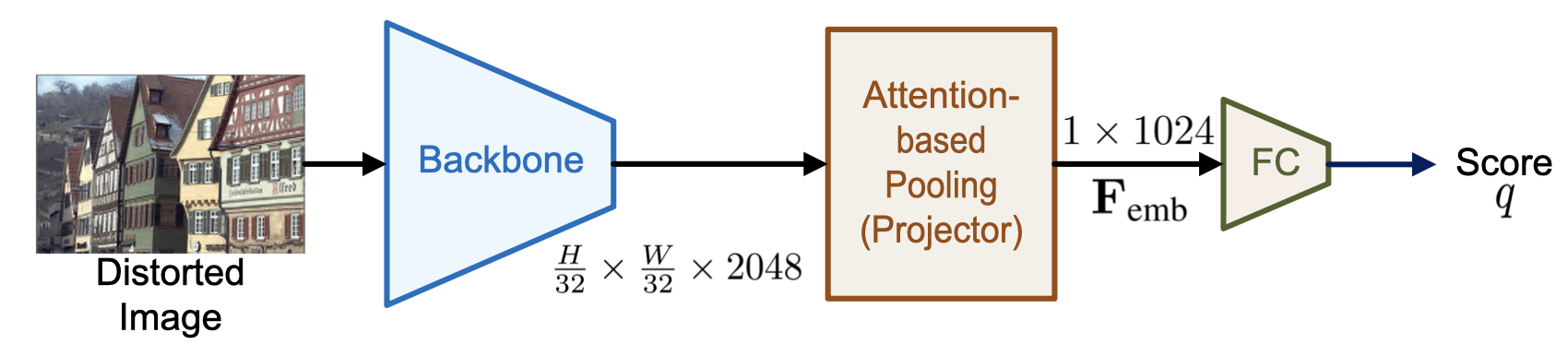}%
    \vspace{-3mm}
    \caption{Inference pipeline of HiRQA. Unlike prior methods, HiRQA requires only a distorted image. Quality-aware features are extracted to predict the quality score $q$. }%
    \label{fig: inference pipeline}
\end{figure}

\vspace{-2mm}
\subsection{Training Strategy and Inference}
Our training strategy is designed to progressively structure the model's understanding of image quality, aligning with the hierarchical quality-aware embedding space we aim to learn. In the initial stage, we optimize the model using two objectives: the ranking loss ($\mathcal{L}_{\text{ranking}}$),  and the image-text contrastive loss ($\mathcal{L}_{\text{align}}$) to learn a consistent global ranking structure. Then we incorporate the embedding distance loss ($\mathcal{L}_{\text{embedding}}$), which is score-dependent and thus introduced only after the first epoch. This loss refines the learned representation by minimizing feature distances between similarly ranked images and maximizing for dissimilar ones, enabling the model to capture fine-grained perceptual differences. The final training objective is defined as:
\begin{equation*}
\mathcal{L}_{\text{final}} = \lambda_{\text{rank}} \mathcal{L}_{\text{ranking}} + \lambda_{\text{align}} \mathcal{L}_{\text{align}} + \lambda_{\text{emb}} \mathcal{L}_{\text{embedding}}
\end{equation*}
where $\lambda_{\text{emb}} = 0$ during the first epoch and is progressively increased in later epochs.

At inference, HiRQA operates solely on a distorted image, without relying on any auxiliary inputs. As shown in Fig.~\ref{fig: inference pipeline}, Quality-aware features extracted by our pretrained backbone are pooled via an attention-based pooling layer into a 1024-dimensional embedding that encodes perceptual quality characteristics. It is then passed to a fully connected \textit{decision layer} to predict the final quality score $q$. 

\begin{table*}[ht]
\centering
\caption{SROCC and PLCC results comparison of HiRQA against SOTA opinion-unaware NR-IQA models on synthetic and authentic datasets are shown. The top three scores are marked in boldfaced \textbf{\color{red}red},  \textbf{\color{blue}blue}, and  \textbf{black} texts, respectively.}
 \vspace{0.5mm}
\resizebox{\linewidth}{!}{
\begin{tabular}{l||cc|cc|cc|cc||cc|cc|cc|cc}
\Xhline{2\arrayrulewidth}
& \multicolumn{8}{c||}{\textbf{Synthetic Datasets}} & \multicolumn{8}{c}{\textbf{Authentic Datasets}}
\\[0.25ex]
\Xhline{\arrayrulewidth}

&
\multicolumn{2}{c|}{LIVE} & \multicolumn{2}{c|}{CSIQ} & \multicolumn{2}{c|}{TID2013} & \multicolumn{2}{c||}{Kadid10k} 
& \multicolumn{2}{c|}{LIVE-C} & \multicolumn{2}{c|}{Koniq10k} & \multicolumn{2}{c|}{LIVE-FB}  & \multicolumn{2}{c}{SPAQ}\\[0.5ex]
\textbf{Model} & \textbf{SROCC} & \textbf{PLCC} & \textbf{SROCC} & \textbf{PLCC} & \textbf{SROCC} & \textbf{PLCC} & \textbf{SROCC} & \textbf{PLCC} & \textbf{SROCC} & \textbf{PLCC} & \textbf{SROCC} & \textbf{PLCC} & \textbf{SROCC} & \textbf{PLCC} & \textbf{SROCC} & \textbf{PLCC} \\[0.4ex]
\Xhline{2\arrayrulewidth}
NIQE~\cite{NIQE}  & 0.903 & \textbf{0.906} & 0.642 & 0.727 & 0.308 & 0.419 & 0.380 & 0.442 & 0.446 & 0.469 & 0.527 & 0.532 & 0.203 & 0.262 & 0.700 & 0.712\\[0.4ex]
IL-NIQE~\cite{ILNIQE}  &0.878 & 0.884 & \textbf{\color{blue}0.848} & \textbf{\color{blue}0.883} & 0.504 & 0.636 & 0.557 & 0.599 & 0.427 & 0.481 & 0.511 & 0.532 & 0.234  & 0.286 & 0.713 & 0.724\\[0.4ex]
CL-MI~\cite{BabuOUIQAAuthentic}  &0.749 & 0.731 & 0.619 & 0.615 & 0.249 & 0.317 & 0.518 & 0.529 & 0.494 & 0.480 & 0.664 & 0.653 & 0.290 & 0.322 & 0.701 & 0.701\\[0.4ex]
MD-FS~\cite{ni2024opinionunawareblindimagequality} &  \textbf{\color{blue}0.926} & \textbf{\color{red}0.930} & 0.797 & 0.829 & 0.554 & 0.657 & 0.608 & 0.634 & 0.427 & 0.467 & 0.739 & 0.746 &  0.360 & 0.407 & 0.805 & 0.812\\[0.4ex]
CONTRIQUE-OU~\cite{madhusudana2021st} & 0.841 & 0.840 & 0.684 & 0.700 & 0.321 & 0.344 & 0.552 & 0.564  & 0.348 & 0.364 & 0.651 & 0.637 & 0.321  & 0.341 & 0.677 & 0.685 \\[0.4ex]
ReIQA-OU~\cite{saha2023re}&  0.781 & 0.779 & 0.713 & 0.725 & 0.284 & 0.339 & 0.516 & 0.534 & 0.379 & 0.288 &  0.580 &  0.568 & 0.350 & 0.341 & 0.613 & 0.616\\[0.4ex]
ARNIQA-OU~\cite{agnolucci2024arniqa} &  0.853 & 0.852 & 0.832 & 0.810 & 0.467 & 0.527 & 0.632 & 0.638  & 0.468 & 0.557 & 0.746 & 0.762 & \textbf{\color{blue}0.428} & 0.468 & 0.788 & 0.797 \\[0.4ex]
CLIP-IQA~\cite{wang2023exploring} & 0.626 & 0.651 & 0.748 & 0.805 & 0.525 & 0.632 & 0.465 & 0.473 & 0.656 & 0.670 & 0.695 &  0.730 & 0.344 & 0.473 & 0.733 & 0.734 \\[0.4ex]
GRepQ-OU~\cite{Srinath_2024_WACV} &   0.714 & 0.698 & 0.740 & 0.742 & 0.423 & 0.565 & 0.416 & 0.463  &  \textbf{0.722} & \textbf{\color{blue}0.766} & 0.776 & 0.798 & 0.289 & 0.423 & 0.802 & 0.805 \\[0.4ex]
QualiCLIP~\cite{agnolucci2024qualityaware} & 0.898 & 0.885 & 0.804 & 0.842 & \textbf{0.651} & \textbf{\color{blue}0.732} & \textbf{0.654} & \textbf{0.664} & \textbf{\color{blue}0.725} & \textbf{\color{red}0.802} & \textbf{\color{red}0.817} & \textbf{\color{red}0.838} & \textbf{\color{red}0.442} & \textbf{\color{red}0.556} & \textbf{0.841} & \textbf{\color{blue}0.851}\\[0.4ex]
\Xhline{2\arrayrulewidth}
\textbf{HiRQA}~(ResNet50) & 0.902  & 0.890 & \textbf{\color{red}0.862} & \textbf{0.874} & \textbf{\color{blue}0.679} & \textbf{0.697} & \textbf{\color{blue}0.761} & \textbf{\color{blue}0.733} & 0.692 & \textbf{0.744} & \textbf{\color{blue}0.802} & \textbf{\color{blue}0.812} & \textbf{0.421} & \textbf{\color{blue}0.490} & \textbf{\color{red}0.859} & \textbf{\color{red}0.860}\\[0.4ex]
\Xhline{2\arrayrulewidth}
\textbf{HiRQA-S}~(ResNet18) & 0.910  & 0.897 & 0.835 & 0.844 & 0.642 & 0.677 & 0.750 & 0.731 & 0.632 & 0.701 & 0.768 & 0.782 & 0.397 & 0.466 & 0.839 & 0.846\\[0.4ex]
\textbf{HiRQA-DinoV3} & \textbf{\color{red}0.939}  & \textbf{\color{blue}0.920} & \textbf{0.838} & \textbf{\color{red}0.885} & \textbf{\color{red}0.731} & \textbf{\color{red}0.742} & \textbf{\color{red}0.793} & \textbf{\color{red}0.759} & \textbf{\color{red}0.727} & 0.737 & 0.772 & 0.791 & 0.401 & 0.478 & 0.839 & 0.846\\[0.4ex]
\textbf{HiRQA-CLIP} & \textbf{0.912}  & 0.902 & 0.834 & 0.853 & 0.676 & 0.688 & 0.748 & 0.733 & 0.702 & 0.734 & \textbf{0.801} & \textbf{0.806 }& 0.417 & \textbf{0.474} & \textbf{\color{blue}0.844} & 0.845\\[0.4ex]
\Xhline{2\arrayrulewidth}
\end{tabular}
}
\label{Tab:Results_NRIQA}
\vspace{-2mm}
\end{table*}

\vspace{-1mm}
\subsection{Implementation Details}\label{imp details}
\vspace{-1mm}
{We implement HiRQA using PyTorch and train on $2$ NVIDIA A100 GPUs for only $3$ epochs. The model is optimized with the AdamW optimizer, starting with a learning rate of $3 e^{-6}$ and a weight decay of $1 e^{-5}$. A Cosine Annealing Learning Rate Scheduler is used, which reduces the Learning rate to a minimum of $8e^{-7}$ over $7000$ iterations.  Additionally, we set the threshold hyperparameters in the loss functions to $T_d=0.2$ and $T_q=0.05$, and keep them fixed across all experiments.}

The training data comprises a diverse set of degraded images generated from the $140$\,K pristine images in the KADIS700k dataset. As per the image degradation model (see Sec.~\ref{sec:idm}), multiple degradations are applied sequentially, up to a maximum $L_{\text{dist}} = 7$ per image. Distortion intensities $d_i$ are sampled continuously by interpolating between discrete levels defined in Kadid10k. This sampling process incorporates a Gaussian offset with zero mean and standard deviation $\sigma_{\text{off}}^2 = 0.3$, for subtle variations in distortion severity. {For each pristine image, we generate five distinct distorted variants using the image degradation model, yielding approximately $700$\,K distorted images for training.}

During training, crops of size $384 \times 384$ pixels are extracted from each image.  During evaluation, we use full-resolution images and the complete dataset without any splitting. However, for SPAQ, due to its high-resolution images, we resize each image such that its shortest side is $768$ pixels while preserving the original aspect ratio.

\section{Experimental Results, Analyses, and Discussion}
We validate the effectiveness of the {HiRQA} framework through a comprehensive set of experiments; our evaluation plan involves the following:
\begin{enumerate}[label={$\arabic*$.},nolistsep,leftmargin=*]
\item \textbf{Zero-shot benchmark evaluation.} We assess how well HiRQA correlates with human visual perception by evaluating it on standard IQA benchmarks. We conduct experiments across eight datasets: LIVE~\cite{sheikh2006statistical}, CSIQ~\cite{larson2010most}, TID2013~\cite{ponomarenko2015image}, Kadid10k~\cite{lin2019kadid}, LIVE Challenge~\cite{ghadiyaram2015massive}, Koniq10k~\cite{hosu2020koniq}, LIVE-FB~\cite{ying2020patches}, and SPAQ~\cite{fang2020cvpr}. We use the standard metrics~\cite{wang2004image,zhai2020perceptual}: Spearman’s rank order correlation coefficient (SROCC) and Pearson’s linear correlation coefficient (PLCC) to evaluate the correlation between the predicted and ground truth scores. PLCC is defined as
\vspace{-2mm}
\begin{equation*}
PLCC=\frac{\sum_{i=1}^N(\hat{q}_i-\Bar{\hat{q}})(q_i-\Bar{q})}{\sqrt{\sum_{i=1}^N(\hat{q}_i-\Bar{\hat{q}})^2}\sqrt{\sum_{i=1}^N(q_i-\Bar{q})^2}}
\end{equation*}
where $N$ is the number of distorted images, $\hat{q}_i$ is the predicted scores, $q_i$ is the true scores; $\Bar{\hat{q}}$ and $\Bar{q}$ are the corresponding means. Besides, SROCC is defined as:
\vspace{-2mm}
\begin{equation*}
SROCC=1-\frac{6\sum^N_{i=1}d^2_i}{N(N^2-1)}
\end{equation*}
where $d_i$ is the pairwise differences between ${q}_i$ and $\hat{q}_i$. 
\item \textbf{Ablation study.} We perform experiments to analyze the contributions of individual loss terms and evaluate our proposed ranking loss for IQA learning.
\item \textbf{Feature  visualization.} We use t-SNE to visualize the learned feature embeddings of HiRQA, offering insights into the perceptual structure captured by the model and the quality-aware clustering behavior.   
\item \textbf{Generalization performance.} To assess the robustness to out-of-distribution distortions, we evaluate its performance on enhancement datasets~\cite{dai2022flare7k,Chen2018Retinex,I-HAZE_2018,nah2017deep} using a density separation criterion~\cite{ramesh2025dgiqadepthguidedfeatureattention} in the supplementary materials.
\end{enumerate}
\vspace{-1mm}

\vspace{1mm}
\noindent


\begin{table*}[t]
\centering
\caption{{Cross-dataset SROCC/PLCC performance of SOTA opinion-aware NR-IQA models trained on Koniq10k and HiRQA is reported. Top-2 results are highlighted in} \textbf{\color{red}red} and \textbf{\color{blue}blue}, {respectively.}}
 \vspace{0.5mm}
\resizebox{0.85\linewidth}{!}{
\begin{tabular}{l||cc|cc|cc|cc|cc}
\Xhline{2\arrayrulewidth}
& \multicolumn{2}{c|}{LIVE-C} & \multicolumn{2}{c|}{SPAQ} &\multicolumn{2}{c|}{LIVE-FB} & \multicolumn{2}{c|}{CSIQ} 
& \multicolumn{2}{c}{Kadid10k}\\[0.5ex]
\textbf{Model} & \textbf{SROCC} & \textbf{PLCC} & \textbf{SROCC} & \textbf{PLCC} & \textbf{SROCC} & \textbf{PLCC} & \textbf{SROCC} & \textbf{PLCC} & \textbf{SROCC} & \textbf{PLCC} \\[0.4ex]
\Xhline{2\arrayrulewidth}
DBCNN~\cite{zhang2018blind} & 0.755 & 0.773 & 0.806 & 0.812 & 0.385 & 0.485 & 0.572 & 0.582  & 0.484 & 0.497  \\[0.4ex]
MUSIQ~\cite{ke2021musiq} & \textbf{\color{blue}0.830} & 0.789 & \textbf{\color{blue}0.863} & \textbf{\color{red}0.868} & \textbf{\color{red}0.467} & \textbf{\color{red}0.565} & 0.710 & 0.771  & 0.556 & 0.575  \\[0.4ex]
ARNIQA~\cite{agnolucci2024arniqa} & 0.724  & 0.759 & 0.805 & 0.808 & 0.410 & 0.487  & 0.595 & 0.571  & 0.620 & 0.611\\[0.4ex]
MANIQA~\cite{yang2022maniqa} & \textbf{\color{red}0.832} & \textbf{\color{red}0.849} & 0.758 & 0.768 & 0.401 & \textbf{\color{blue}0.512} & 0.627 & 0.623 & 0.465 & 0.499  \\[0.4ex]
CLIP-IQA+~\cite{wang2023exploring}&  0.805 & \textbf{\color{blue}0.832} & \textbf{\textcolor{red}{0.864}} & \textbf{\textcolor{blue}{0.866}} & 0.316 & 0.427 & \textbf{\color{blue}0.719} & \textbf{\color{blue}0.772} & \textbf{\color{blue}0.654} & \textbf{\color{blue}0.653} \\[0.4ex]

\Xhline{2\arrayrulewidth}
\textbf{HiRQA~(ours)} & 0.692  & 0.744 & 0.859 & 0.860 & \textbf{\color{blue}0.421} & 0.490 & \textbf{\color{red}0.862} & \textbf{\color{red}0.874} & \textbf{\color{red}0.761} & \textbf{\color{red}0.733} \\[0.4ex]
\Xhline{2\arrayrulewidth}
\end{tabular}
}
\label{Tab:Results_opinion_aware_cross}
\vspace{-3mm}
\end{table*}

\vspace{1mm}

\subsection{Zero-Shot Benchmarks}
\vspace{-1mm}
We compare HiRQA with several SOTA OU-NRIQA models, including traditional statistical approaches~\cite{NIQE,ILNIQE}), self-supervised models~\cite{BabuOUIQAAuthentic,saha2023re,madhusudana2021st,agnolucci2024arniqa}, and VLM frameworks~\cite{agnolucci2024qualityaware,Srinath_2024_WACV,wang2023exploring}. As noted in~\cite{agnolucci2024qualityaware}, self-supervised models rely on a NIQE-style framework that measures the distance between features of distorted images and those from a corpus of pristine references to compute the quality score.

\Rebute{ As shown in Table~\ref{Tab:Results_NRIQA}, HiRQA delivers consistently competitive performance across all eight IQA benchmark datasets, achieving {top-three rankings} in both SROCC and PLCC across all test sets. It attains the \underline{highest scores} on TID2013, Kadid10k, CSIQ, and SPAQ, and offers competitive scores on authentic datasets such as LIVE-C, Koniq10k, and LIVE-FB. Notably, this performance is achieved using a single ResNet-50 backbone \underline{without relying on} auxiliary modalities or reference images during inference.}

\Rebute{Unlike methods such as GRepQ, which utilize a dual-encoder architecture, or ARNIQA, which operates on both full and half-resolution images, HiRQA achieves better performance with a \underline{single image encoder} operating on full-resolution images, avoiding multi-resolution inputs or multi-branch networks. While models like QualiCLIP and GRepQ fine-tune CLIP encoders, that are already known to exhibit strong zero-shot capabilities for IQA~\cite{wang2023exploring},  HiRQA is trained using a ResNet backbone pretrained on ImageNet and leverages only a text-encoder-based loss during training. Our method \underline{eliminates the dependency} on large VLMs, yet achieves competitive or superior performance against these methods on most benchmark datasets.} 

{HiRQA also offers \underline{architectural flexibility} over existing CLIP-based methods. For instance, methods such as QualiCLIP and GRepQ are inherently constrained by the CLIP architecture, offering limited flexibility in backbone since the vision encoder must remain compatible with CLIP's pretraining. In contrast, HiRQA can be used with any standard backbone architecture. We demonstrate this by \textbf{HiRQA-S}, which uses a lightweight ResNet-18 as its backbone and achieves an inference time of $3.5$~ms per image. As presented in Table~\ref{Tab:Results_NRIQA}, HiRQA-S attains competitive performance even with a much smaller backbone, demonstrating the efficiency of our design.} 

\Rebute{ While HiRQA's performance on LIVE-FB is more limited, this reflects a dataset-level challenge where all evaluated methods struggle. A key contributing factor is the extreme resolution variance in LIVE-FB~\cite{ying2020patches}, where shorter-side resolutions range from 160px to 700px. A resolution-stratified analysis shows that images below our training resolution of $384$px (52.9\% of the dataset) achieve SROCC of $0.401$, while images above $384$px achieve SROCC of $0.436$, suggesting resolution mismatch between the dataset and training as a contributing factor to the performance gap.}

\vspace{1mm}
\noindent \Rebute{\textbf{Backbone variants.} The default HiRQA model uses a ResNet-50 backbone pretrained on ImageNet, consistent with the majority of compared self-supervised methods. To further assess architectural generality, Table~\ref{Tab:Results_NRIQA} reports HiRQA performance with several pretrained backbones: ResNet50 (default), ResNet18 (HiRQA-S), DINOv3-based ConvNeXt-Base encoder (HiRQA-DINOv3)~\cite{simeoni2025dinov3}, and CLIP-initialized ResNet-50 (HiRQA-CLIP)~\cite{radford2021learningtransferablevisualmodels}. HiRQA-DINOv3 performs best on the synthetic benchmarks and also improves SROCC on LIVE-C. Across datasets, ResNet-50 offers the most consistent trade-off between synthetic and authentic performance. }

For HiRQA-CLIP, fine-tuning appears to weaken CLIP’s original vision–language alignment priors, resulting in performance close to the ImageNet-pretrained ResNet50. More generally, attention-based backbones tend to exhibit weaker inductive bias than CNNs, and prior OU-IQA works report that transformer backbones often underperform convolutional encoders under comparable settings~\cite{wang2023exploring,agnolucci2024qualityaware}. We hypothesize that this reduced inductive bias increases sensitivity during self-supervised adaptation, which may contribute to the slightly less balanced performance of the DINOv3 variant relative to ResNet50. Additionally, in the supplementary material, we have shown that HiRQA outperforms both CLIP-IQA and QualiCLIP when all methods use the same ViT-B/32 backbone.

\vspace{2mm}
\noindent\textbf{{Comparison with opinion-aware NR-IQA models.}} {Table~\ref{Tab:Results_opinion_aware_cross} reports cross-dataset performance of SOTA opinion-aware NR-IQA models trained on Koniq10k against HiRQA. Overall, HiRQA achieves performance that is similar to the cross-dataset performance of supervised opinion-aware models, and is competitive with, or better than, several of these baselines despite not using subjective labels during training. We attribute this behavior to the fact that subjective labels can introduce dataset-specific bias; models trained on a particular dataset may overfit its content and opinion distribution, yielding strong performance on some datasets but noticeably weaker transfer to others. In contrast, HiRQA learns a transferable quality representation that reduces reliance on subjective labels and generalizes across domains.}

\vspace{2mm}
\noindent \Rebute{\textbf{Out-of-distribution generalization.} Apart from the standard benchmark evaluation, to further verify the out-of-distribution generalization ability of HiRQA, we evaluate it on three additional unseen domains: AI-generated images~\cite{li2023agiqa}, underwater images~\cite{hou2023uid2021}, and image restoration~\cite{sheng2026fine}. As shown in Table~\ref{Tab:ood_perfromance}, HiRQA achieves competitive or superior performance compared to QualiCLIP across all three domains. Notably, on UID2021, HiRQA significantly outperforms QualiCLIP, despite being trained exclusively on synthetic distortions from KADIS-700K with no exposure to underwater images or underwater-specific degradations.}
\begin{table}[t]
\caption{\Rebute{Zero-shot performance of HiRQA against QualiCLIP on unseen out-of-distribution benchmarks spanning underwater, AI-generated, and image restoration domains. Top results are highlighted in} \textbf{bold}.}
\label{Tab:ood_perfromance}
\centering
\begin{tabular*}{\columnwidth}{@{\extracolsep{\fill}} l|cc|cc @{}}
\Xhline{2\arrayrulewidth}
Method $\rightarrow$ & \multicolumn{2}{c|}{QualiCLIP} & \multicolumn{2}{c}{HiRQA} \\
\Xhline{2\arrayrulewidth}
Dataset $\downarrow$ & SROCC & PLCC & SROCC & PLCC \\
\Xhline{2\arrayrulewidth}
AGIQA-3K~\cite{li2023agiqa}       & \textbf{0.651} & 0.704 & 0.643 & \textbf{0.710}  \\
UID2021~\cite{hou2023uid2021}           & 0.265 & 0.274 & \textbf{0.417} & \textbf{0.421}  \\
FGRestore~\cite{sheng2026fine} & 0.360 & 0.343 & \textbf{0.447} & \textbf{0.445} \\
\Xhline{2\arrayrulewidth}
\end{tabular*}%
\vspace{-6mm}
\end{table}

\begin{table}[b]
\centering
\caption{{ Model size and inference runtime comparison of HiRQA against other NR-IQA methods on Koniq10k (1024×768). Number of parameters(\# Params) are reported in millions (M). Runtime is measured in milliseconds per image.}}
\label{Tab:inference_params}
\begin{tabular*}{\columnwidth}{@{\extracolsep{\fill}} l|c|c @{}}
\Xhline{2\arrayrulewidth}
\textbf{Model} & \textbf{\# Params} & \textbf{Runtime (ms)} \\
\Xhline{2\arrayrulewidth}
ARNIQA~\cite{agnolucci2024arniqa}     & 28.0~M   & 5.0 \\
MANIQA~\cite{yang2022maniqa}    & 135.7~M   &  527.2  \\
CLIP-IQA~\cite{wang2023exploring} & 38.3~M   & 9.8 \\
GRepQ~\cite{Srinath_2024_WACV}      & 50.0~M & 22.5 \\
QualiCLIP~\cite{agnolucci2024qualityaware}  & 38.3~M   & 9.9 \\
\Xhline{2\arrayrulewidth}
\textbf{HiRQA}      & \textbf{38.2~M}& \textbf{9.2} \\
\textbf{HiRQA-S}    & \textbf{12.5~M} & \textbf{3.5} \\
\Xhline{2\arrayrulewidth}
\end{tabular*}
\vspace{-2mm}
\end{table}

\vspace{2mm}
{
\noindent \textbf{Model Size and Inference Runtime.} Table~\ref{Tab:inference_params} reports the computational footprint of HiRQA compared to SOTA NR-IQA baselines spanning both opinion-aware and opinion-unaware settings, in terms of model size (\# Params) and inference latency (Runtime). Runtime is measured in milliseconds per image under identical inference settings on the Koniq10k benchmark, using a single NVIDIA A100 GPU. Overall, HiRQA provides a favorable efficiency–accuracy trade-off, using $38.2M$ parameters with a $9.2$~ms per image runtime, while the lightweight HiRQA-S further reduces the footprint to $12.5M$ parameters and $3.5$~ms per image.}

\vspace{-1mm}
\subsection{Ablation Experiments}\label{sec:ablation}
\vspace{-1mm}
{
We conduct comprehensive ablation studies to quantify the contribution of key components in the HiRQA framework. In addition to analyzing the ranking formulation, we compare our proposed higher-order supervision (Pair-of-Pairs) against standard pairwise objectives (RankNet and margin loss), and study the impact of design choices in the image degradation model, prompt formulation, and training strategy. The main results are summarized in Tables~\ref{Tab:Ablation_Loss}-\ref{Tab:additional_ablation_analysis}, and an extended set of ablation for Table~\ref{Tab:additional_ablation_analysis} is provided in the supplementary material. 
}

\vspace{1mm}
\noindent
\textbf{Contributions of Loss Terms.} We conduct an ablation study on the Koniq10k dataset by selectively enabling the ranking loss $\mathcal{L}_{\text{ranking}}$, the embedding distance loss $\mathcal{L}_{\text{embedding}}$, and the text-image alignment loss $\mathcal{L}_{\text{align}}$. As the results in Table~\ref{Tab:Ablation_Loss} demonstrate, using only $\mathcal{L}_{\text{ranking}}$ serves as a baseline for relative ordering among distorted images. This improves the model's ability to rank image quality, as reflected in higher SROCC. However, it lacks explicit supervision for the magnitude of the predicted scores, leading to slightly lower PLCC scores. 
\begin{table}[t]
\vspace{-2mm}
\centering
\footnotesize
\caption{Ablation results on Koniq10K dataset are shown for four configurations of the proposed HiRQA model.}
\vspace{-2mm}
\begin{tabular}{c|ccc|cc}
\Xhline{2\arrayrulewidth}
Cfg & $\mathcal{L}_{\text{ranking}}$ & $\mathcal{L}_{\text{embedding}}$ &$\mathcal{L}_{\text{align}}$ & SROCC & PLCC \\ 
\Xhline{2\arrayrulewidth}
\#1 & \checkmark &  &  & 0.751 & 0.730 \\ 
\#2 &\checkmark  & \checkmark &  & 0.765 & 0.752 \\ 
\#3 & \checkmark &  & \checkmark  & 0.782 & 0.771 \\ 
\#4 & \checkmark & \checkmark & \checkmark & \textbf{0.802} & \textbf{0.812} \\ 
\Xhline{2\arrayrulewidth}
\end{tabular}%
\vspace{-5mm}
\label{Tab:Ablation_Loss}
\end{table}

To structure the learned embedding space, we incorporate the embedding loss $\mathcal{L}_{\text{embedding}}$, which supervises the differences in predicted quality scores based on embedding similarity. This encourages the predicted scores to better reflect perceptual differences, leading to a $1.9\%$ ($3.01\%$) improvement in SROCC (PLCC) over the baseline. Similarly, introducing the text-image alignment loss $\mathcal{L}_{\text{align}}$ guides the model using weak semantic supervision, improving SROCC (PLCC) by $4.1\%$ ($5.6\%$). This shows that aligning image features with distortion-aware text prompts provides additional structure that benefits both ranking and magnitude of the predicted scores. When all three loss terms are combined, HiRQA achieves the best performance with a $6.8\% $($11.1\%$) increase in SROCC (PLCC) compared to the baselines. 


\vspace{2mm}
\noindent
\textbf{Ranking Loss Variants.} We also compare three ranking loss strategies to understand their effect on learning quality-aware representations; see Table~\ref{Tab:ablation_ranking_supervision}. The standard pairwise RankNet loss encourages correct ordering between image pairs but lacks global structural context, resulting in limited performance. A margin-based loss, which enforces both correct ranking and a minimum perceptual separation between predicted scores, yields stronger supervision and leads to notable gains, improving SROCC (PLCC) by 14.9\% (23.7\%) on LIVE-C and $7.9\%$ ($9.3\%$) on Koniq10K, respectively. 


\begin{table}[t]
\centering
\caption{Ablation on ranking formulation in HiRQA; results are shown for LIVE-C and Koniq10k.}
\label{Tab:ablation_ranking_supervision}
\begin{tabular*}{\columnwidth}{@{\extracolsep{\fill}} l|cc|cc @{}}
\Xhline{2\arrayrulewidth}
Test on $\rightarrow$ & \multicolumn{2}{c|}{LIVE-C} & \multicolumn{2}{c}{Koniq10k} \\
\Xhline{2\arrayrulewidth}
Ranking Loss & SROCC & PLCC & SROCC & PLCC \\
\Xhline{2\arrayrulewidth}
Pairwise RankNet       & 0.557 & 0.568 & 0.701 & 0.743  \\
Margin Loss            & 0.640 & 0.703 & 0.757 & 0.775  \\
\textbf{Pair-of-Pairs} & \textbf{0.692} & \textbf{0.744} & \textbf{0.802} & \textbf{0.812} \\
\Xhline{2\arrayrulewidth}
\end{tabular*}%
\vspace{-6mm}
\end{table}



\Rebute{Unlike RankNet, which mainly enforces pairwise ordering, and margin-based ranking, which imposes a fixed separation, pair-of-pairs provides an adaptive gap constraint based on relative distortion differences. As shown in Table~\ref{Tab:ablation_ranking_supervision}, it consistently outperforms both baselines on LIVE-C and KonIQ-10k, improving over margin-based ranking by $8.1\%$($5.8\%$) in SROCC(PLCC) on LIVE-C and $5.9\%$($4.8\%$) on KonIQ-10k. These results show that adaptive higher-order ranking improves perceptual alignment beyond both pairwise ordering and fixed-margin separation.}

\vspace{1mm}
\noindent\textbf{Image Degradation Model.}
{We conduct an ablation study of two components of our image degradation model: (a) the distortion level sampling strategy and (b) the degradation hyperparameters ($L_{\mathrm{dist}}$, $\sigma_{\mathrm{off}}^2$).  In Table~\ref{Tab:additional_ablation_analysis}(a), we compare the continuous distortion variance introduced in Section~\ref{sec:idm} against the discrete variant used by ARNIQA~\cite{agnolucci2024arniqa}. Replacing discrete levels with \underline{interpolated continuous levels} improves SROCC(PLCC) by $0.75\%$($2.01\%$), indicating that exposing the model to fine-grained perceptual quality variations strengthens the ranking supervision. In Table~\ref{Tab:additional_ablation_analysis}(b), we vary the degradation hyperparameters and observe that increasing the maximum number of sequential distortions generally improves performance. Among the tested settings, $L_{\mathrm{dist}}{=}7$ with $\sigma_{\mathrm{off}}^2{=}0.3$ yields the best overall performance.}
\vspace{1mm}
\begin{table}[t]
\centering
\caption{{Additional ablation study on (a) Distortion variance (discrete vs.\ continuous interpolation), (b) Image degradation model hyperparameters ($L_{\mathrm{dist}}, \sigma_{\mathrm{off}}^2$),(c) Prompt components, (d) Training strategy for $\lambda_{\mathrm{emb}}$ at epoch~$1$, (e) Distortion aggregation strategy, and (f) Ranking loss thresholds $T_q$ and $T_d$. Results are reported as SROCC/PLCC on Koniq10k. The HiRQA setting is highlighted in} \textbf{bold}.}
\label{Tab:additional_ablation_analysis}
\begin{tabular*}{\columnwidth}{@{\extracolsep{\fill}} c|c||c|c @{}}
\Xhline{2\arrayrulewidth}
\multicolumn{2}{c||}{Test on $\rightarrow$}  & \multicolumn{2}{c}{Koniq10k} \\
\Xhline{2\arrayrulewidth}
\multicolumn{2}{c||}{Configuration}  & SROCC & PLCC \\
\Xhline{2\arrayrulewidth}

\multicolumn{4}{c}{(a) Distortion variance.} \\
\Xhline{2\arrayrulewidth}
\multicolumn{2}{c||}{Discrete} & 0.796 & 0.796  \\
\multicolumn{2}{c||}{\textbf{Continuous}} & \textbf{0.802 } & \textbf{0.812 }  \\
\Xhline{2\arrayrulewidth}
\multicolumn{4}{c}{(b) Image degradation model hyperparameters.} \\[0.2ex]
\Xhline{2\arrayrulewidth} 
$L_{\mathrm{\text{dist}}}\phantom{^{2}_{\text{off}}}$ & $\sigma_{\mathrm{off}}^2$ & \multicolumn{2}{c}{ } \\[0.2ex]
\Xhline{2\arrayrulewidth}
3  \phantom{0.0} & 0.3  & 0.764 &   0.787\\
5 \phantom{0.0}  & 0.3  & 0.785 &  0.793 \\
\textbf{7} \phantom{0.0}  & \textbf{0.3}  & \textbf{0.802 } & \textbf{0.812}  \\
7  \phantom{0.0} & 0.2  & 0.796 &  0.808 \\
7 \phantom{0.0}  & 0.1  & 0.791 & 0.799   \\
\Xhline{2\arrayrulewidth}

\multicolumn{4}{c}{(c) Prompt components.} \\
\Xhline{2\arrayrulewidth}
\multicolumn{2}{c||}{No prompt} & 0.765 & 0.752  \\
\multicolumn{2}{c||}{Only semantic caption} & 0.779 & 0.782  \\
\multicolumn{2}{c||}{Only distortion clause} & 0.740  &  0.766 \\
\multicolumn{2}{c||}{\textbf{Full prompt}} &  \textbf{0.802 } & \textbf{0.812 }  \\
\Xhline{2\arrayrulewidth}
\multicolumn{4}{c}{(d) Training Strategy for $\lambda_{emb}$ at epoch~$1$.} \\
\Xhline{2\arrayrulewidth}
\multicolumn{2}{c||}{$\lambda_{emb} \neq 0$} & 0.799 &  0.796 \\
\multicolumn{2}{c||}{$\mathbf{\lambda_{emb} = 0}$} & \textbf{0.802 } & \textbf{0.812 }  \\
\Xhline{2\arrayrulewidth}
\multicolumn{4}{c}{(e) Distortion Aggregation Strategy.} \\
\Xhline{2\arrayrulewidth}
\multicolumn{2}{c||}{Mean} & 0.661 &  0.640 \\
\multicolumn{2}{c||}{Maximum} & \textbf{0.802 } & \textbf{0.812 }  \\
\Xhline{2\arrayrulewidth}
\multicolumn{4}{c}{(f) Ranking Loss Thresholds, $T_d$ and $T_q$.} \\[0.2ex]
\Xhline{2\arrayrulewidth} 
$T_d$ & $T_q$ & \multicolumn{2}{c}{ } \\[0.2ex]
\Xhline{2\arrayrulewidth}
0   & 0.05 & 0.793 &   0.808\\
1   & 0.05  & 0.792 &  0.778 \\
\textbf{0.2}   & \textbf{0.05}  & \textbf{0.802 } & \textbf{0.812}  \\
0.2   & 0.1  & 0.798 &  0.794 \\
0.2   & 0.5  & 0.789 & 0.790   \\
\Xhline{2\arrayrulewidth}
\end{tabular*}%
\vspace{-5mm}
\end{table}

\noindent\Rebute{\textbf{Prompt component.} As described in Section~\ref{sec:img-text-align}, we adopt a structured prompt containing a distortion clause and a semantic caption. Table~\ref{Tab:additional_ablation_analysis}(c) analyzes their contributions. Removing the prompt degrades performance, while using only the semantic caption outperforms the distortion clause alone. The weaker performance of the distortion-clause-only setting is likely due to its coarse text supervision, since many images share identical or near-identical text descriptions from a fixed set of distortion types and five severity adjectives. BLIP-generated captions provide image-specific semantic grounding, making the alignment loss more discriminative. Thus, semantic captions act as content-aware anchors while distortion clauses provide coarse degradation-aware structure, and their combination yields the best overall performance. }
\vspace{2mm}

\noindent\textbf{Training Strategy.} {In Table~\ref{Tab:additional_ablation_analysis}(d), we analyze the strategy of introducing the embedding-distance consistency ($\mathcal{L}_{\text{embedding}}$) only after the model has first learned the ranking structure. Specifically, setting $\lambda_{\mathrm{emb}}=0$ for epoch~$1$ stabilizes early optimization and improves the final results by $0.3\%$($2.01\%$) in SROCC(PLCC).}

\vspace{2mm}
\noindent \Rebute{\textbf{Distortion Aggregation Strategy.} Table~\ref{Tab:additional_ablation_analysis}(e) compares two strategies for computing the overall severity of multi-distortion images: mean and maximum distortion level. The maximum aggregation significantly outperforms mean aggregation, improving SROCC(PLCC) by $21.3\%$ ($26.9\%$) respectively. Mean aggregation underestimates severity when one distortion clearly dominates, introducing inconsistent ranking labels that degrade learning. This validates our design choice of maximum aggregation as an effective approximation for multi-distortion severity.}

\vspace{2mm}
\noindent \Rebute{\textbf{Ranking Loss Thresholds.} Table~\ref{Tab:additional_ablation_analysis}(f) analyzes the sensitivity of HiRQA to $T_d$ and $T_q$, which filter out pairs with near-identical distortion levels and score differences respectively. The results show that HiRQA is robust to threshold choices, with all configurations achieving competitive performance within a narrow range. Setting $T_d=0$ includes noisy pairs while $T_d=1$ discards too many valid pairs, and larger $T_q$ values restrict the embedding loss to only extreme quality gaps, reducing fine-grained supervision. $T_d=0.2$ and $T_q=0.05$ provide the optimal trade-off and are kept fixed across all experiments.}


\vspace{-1mm}
\subsection{Feature Visualization}
\begin{figure}[t]
\centering
\includegraphics[width=0.98\linewidth]{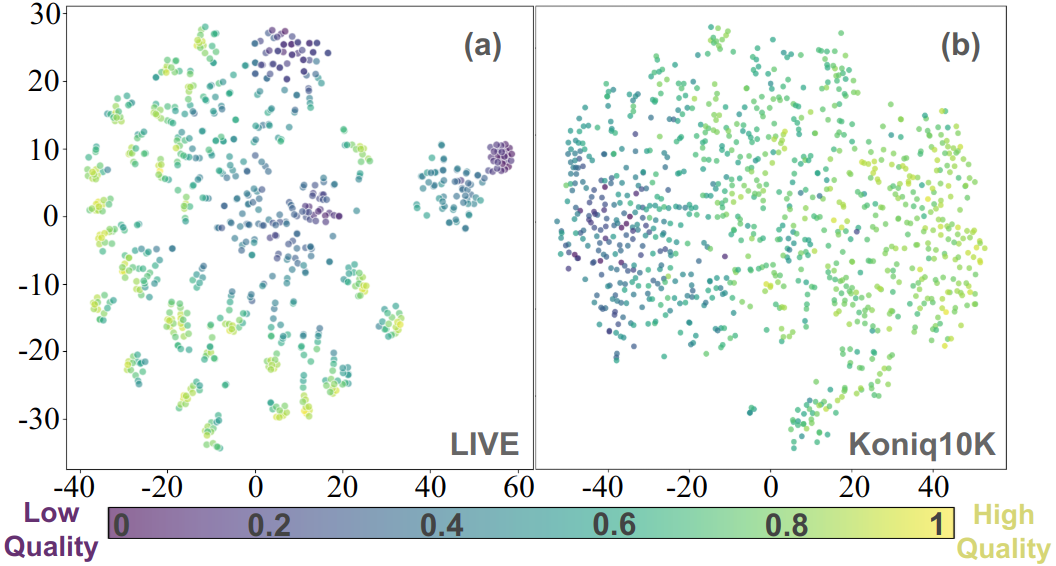}%
\caption{\Rebute{t-SNE visualization of HiRQA's learned embeddings on (\textbf{a}) LIVE and (\textbf{b}) Koniq10K. The results highlight HiRQA’s ability to learn a perceptually structured and separable feature space.}}
\label{fig:TSNE}
\end{figure}

\vspace{-1mm}
We now investigate the hierarchical feature space we aimed to learn. We use t-distributed stochastic neighbor embedding (t-SNE) to visualize the high-dimensional feature representations of HiRQA in a 2D space. In Fig.~\ref{fig:TSNE}, we present t-SNE visualizations of distorted image embeddings from two datasets: LIVE and Koniq10K. For the LIVE dataset (Fig.~\ref{fig:TSNE}a), which contains synthetically distorted images, we visualize all distorted images colored by their ground truth Mean Opinion Scores (MOS). A clear quality gradient is observed where high- and low-quality images \underline{cluster separately}. This confirms that HiRQA learns a structured, quality-aware feature space, where perceptually similar images are placed closer together.

For Koniq10K (Fig.~\ref{fig:TSNE}b), which comprises authentically distorted images, due to its large scale, we randomly sample $500$ high and low-quality images based on their ground truth MOS scores. A clear quality gradient is observed in the learned space, with high- and low-quality images occupying distinct regions and a \underline{smooth transition} across the quality spectrum. This highlights HiRQA's ability to generalize to complex authentic distortions. Together, these results confirm that HiRQA captures meaningful, hierarchical quality representations across both synthetic and authentic domains, aligning well with human visual perception.

\vspace{-2mm}
\subsection{Discussion: Generalization, Limitations, and Future Work}\label{sec:gen}

\begin{figure}[t]
\centering
\includegraphics[width=0.98\linewidth]{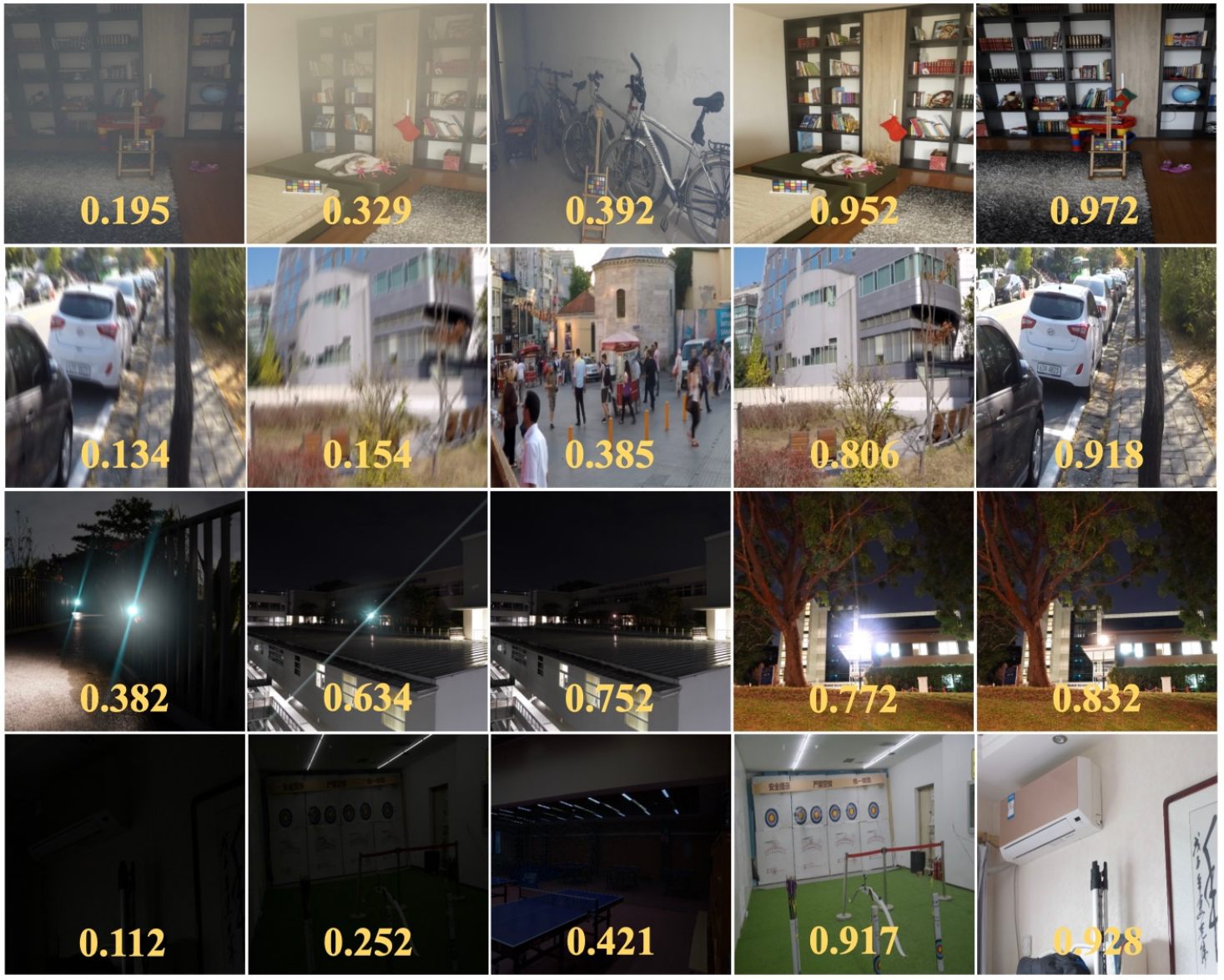} \\     
\caption{Predicted quality scores by HiRQA for haze (row 1), motion blur (row 2), flare (row 3), and low-light (row 4) distortions. In each row, distortion severity increases from right to left; as shown, the corresponding quality scores of HiRQA also decrease in that order -- validating its effectiveness on representative out-of-distribution examples.} 
\label{fig:hirqa_examples}
\end{figure}

{Fig.~\ref{fig:hirqa_examples} presents representative samples from enhancement datasets~\cite{I-HAZE_2018,dai2022flare7k,nah2017deep,Chen2018Retinex}, with HiRQA’s predicted scores \underline{consistently decreasing} as \underline{distortion severity increases} in unseen images across haze, motion blur, lens flare, and low-light~conditions. Beyond the standard benchmark evaluation, we include an additional generalization analysis \textbf{in the supplementary material} using the density separation criterion~\cite{ramesh2025dgiqadepthguidedfeatureattention} on paired real-world distortion datasets. We further provide additional qualitative examples illustrating HiRQA’s predicted quality ordering across representative authentic degradations.}

\vspace{1mm}
\noindent
{
\textbf{Limitations and Future Work.} While HiRQA demonstrates strong performance in OU-NRIQA, the training relies on a simplifying assumption. For images with multiple distortions, we assume that the overall distortion severity corresponds to the maximum distortion level among the applied transformations. This may not always reflect perceptual dominance, especially when multiple moderate distortions jointly degrade quality. As a result, the current formulation does not fully capture the cumulative effect of multiple distortions. Second, our current degradation model is a controllable but simplified approximation and does not fully capture the physical coupling present in real imaging pipelines. Although this choice enables scalable self-supervised ranking supervision, it may contribute to the remaining synthetic-to-authentic gap.}

{
Moreover, like other OU-NRIQA models, HiRQA still lags behind some fully supervised counterparts. Promising future directions include: (i) designing distortion-aware aggregation schemes that jointly account for maximum severity, the number, and type of distortions, and (ii) developing more realistic yet controllable synthesis pipelines to further improve transfer to authentic distortions. Such a refinement could enable more accurate and perceptually consistent quality predictions, particularly in images affected by complex or compound distortions. }


\vspace{-2mm}
\section{Conclusion}
\vspace{-1mm}
\Rebute{ We presented HiRQA, a novel opinion-unaware NR-IQA framework that learns a hierarchical, quality-aware embedding space using synthetic distortions and relative ranking signals. Unlike prior methods that require pristine references or pre-trained vision-language encoders, HiRQA operates using only a single distorted image and image encoder at inference. Our approach combines a higher-order pair-of-pairs ranking loss with embedding distance alignment and semantic guidance via structured text prompts during training. This self-supervised approach enables HiRQA to learn perceptually meaningful features without any subjective labels. Through extensive experiments on both synthetic and authentic IQA datasets, HiRQA consistently achieves SOTA performance on synthetic benchmarks and remains competitive on authentic datasets. It learns a perceptually structured embedding space, as shown through t-SNE visualizations. We further demonstrate the scalability of our framework by substituting the backbone with a light-weight variant (HiRQA-S), which achieves competitive performance while enabling faster inference.} 

\vspace{-1mm}
\section*{Funding Information}
\vspace{-1mm}
This work is supported in part by the National Science Foundation (NSF) award \#$2330416$ and the University of Florida research grant \#$132763$. 

\vspace{-1mm}
\section*{Data availability statement}
\vspace{-1mm}

The learning pipeline, experimental data, and analysis scripts developed as part of this work are made openly available to the broader research community. The repository also contains documentation to facilitate reproducibility. The access is provided online at: \url{https://github.com/uf-robopi/HiRQA}.
{
\bibliographystyle{ieee_fullname}
\bibliography{refs}
}

\end{document}